\documentclass[letterpaper]{article} 
\usepackage{aaai23}  
\usepackage{times}  
\usepackage{helvet}  
\usepackage{courier}  
\usepackage[hyphens]{url}  
\usepackage{graphicx} 
\usepackage{natbib}  
\usepackage{caption} 
\usepackage{algorithm}
\usepackage{algorithmicx}

\usepackage{newfloat}
\usepackage{listings}
\DeclareCaptionStyle{ruled}{labelfont=normalfont,labelsep=colon,strut=off} 
    \floatstyle{ruled}
    \newfloat{listing}{tb}{lst}{}
    \floatname{listing}{Listing}
\usepackage{booktabs}  
\usepackage{amsmath}
\usepackage{amssymb}
\usepackage{interval}
\usepackage{xcolor}
\usepackage{subcaption}
\usepackage{multirow}
\usepackage{enumitem}
\usepackage{wrapfig}
\usepackage{bbm}
\usepackage{tabularx,lipsum}

\usepackage{algpseudocode}
\usepackage[capitalise]{cleveref}
\graphicspath{{./}{pics/}}
\newcommand{\bA}{\mathbf{A}}
\newcommand{\bs}{\mathbf{s}}
\newcommand{\brho}{\boldsymbol{\rho}}
\newcommand{\btau}{\boldsymbol{\tau}}

\newcommand{\note}[2]{\textcolor{#1}{#2}}

\newcommand{\zhang}[1]{\note{black}{#1}}
\newcommand*\samethanks[1][\value{footnote}]{\footnotemark[#1]}

\title{Multi-Action Dialog Policy Learning from Logged User Feedback}
\author {
  Shuo Zhang\textsuperscript{\rm 1},
  Junzhou Zhao\textsuperscript{\rm 1}\thanks{ Corresponding Author },
  Pinghui Wang\textsuperscript{\rm 1}\samethanks[1], 
  Tianxiang Wang\textsuperscript{\rm 1},
  \\ 
  Zi Liang\textsuperscript{\rm 1},
  Jing Tao\textsuperscript{\rm 1},
  Yi Huang\textsuperscript{\rm 2},
  Junlan Feng\textsuperscript{\rm 2} 
  \\
}
\affiliations {
    \textsuperscript{\rm 1} MOE KLINNS Lab, Xi'an Jiaotong University, Xi'an 710049, P. R. China \\
    \textsuperscript{\rm 2} JIUTIAN Team, China Mobile Research\\
    \{zs412082986, liangzid\}@stu.xjtu.edu.cn, \{junzhou.zhao, phwang, jtao\}@mail.xjtu.edu.cn,\\ 2673021501@qq.com, \{huangyi, fengjunlan\}@chinamobile.com\\
}
\begin{document}

\maketitle

\begin{abstract}
  Multi-action dialog policy, which generates multiple atomic dialog actions per turn, has been widely applied in task-oriented dialog systems to provide expressive and efficient system responses. Existing policy models usually imitate action combinations from the labeled multi-action dialog examples. Due to data limitations, they generalize poorly toward unseen dialog flows. While reinforcement learning-based methods are proposed to incorporate the service ratings from real users and user simulators as external supervision signals, they suffer from sparse and less credible dialog-level rewards. To cope with this problem, we explore to improve multi-action dialog policy learning with explicit and implicit turn-level user feedback received for historical predictions (i.e., logged user feedback) that are cost-efficient to collect and faithful to real-world scenarios. The task is challenging since the logged user feedback provides only partial label feedback limited to the particular historical dialog actions predicted by the agent. To fully exploit such feedback information, we propose BanditMatch, which addresses the task from a feedback-enhanced semi-supervised learning perspective with a hybrid objective of semi-supervised learning and bandit learning.
  \zhang{BanditMatch integrates pseudo-labeling methods to better explore the action space through constructing full label feedback.} 
  Extensive experiments show that our BanditMatch outperforms the state-of-the-art methods by generating more concise and informative responses. The source code and the appendix of this paper can be obtained from \url{https://github.com/ShuoZhangXJTU/BanditMatch}. 
\end{abstract}

\section{Introduction}\label{sec:intro}
Dialog policy, which takes the response by deciding the next system action, is critical in task-oriented dialog systems as it determines service quality~\cite{zhang2020recent}.
To further strengthen the expressive power of the agent, recent works have investigated multi-action dialog policy learning (MADPL), which simultaneously generates multiple actions as the current system response~\cite{shu2019modeling,jhunjhunwala2020multi,zhang2022ijcai}.
MADPL is usually formulated as a multi-label classification problem and thus solved by supervised learning (SL) based methods that imitate action combinations from labeled dialog examples~\cite{shu2019modeling,jhunjhunwala2020multi,li2020rethinking}.
However, the complex action combinations can exponentially enlarge the output space. SL-based models trained on limited training corpus alone do not generalize well to real-world dialog flows~\cite{jhunjhunwala2020multi}.

\begin{figure}[tp]
  \includegraphics[width=\linewidth]{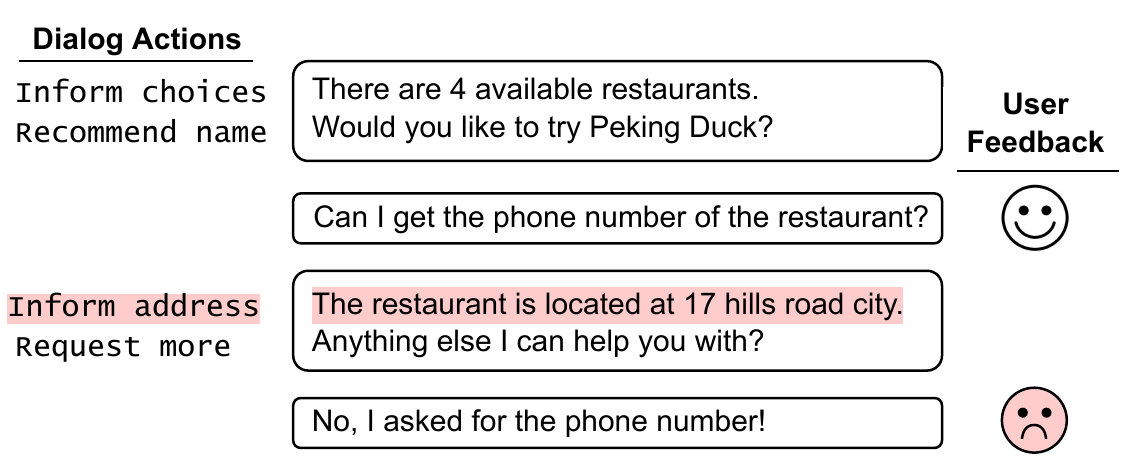}
  \centering
  \caption{Example dialog fragment under multi-action dialog policy. Failing to generate a proper dialog action combination  can trigger negative user feedback.}
  \label{fig:casefig}
\end{figure}

Some reinforcement learning (RL) methods are proposed to address the above issue by learning through interacting with real users while suffer from sparse and unstable reward signals~\cite{takanobu2019guided,li2020guided}.
The reward, which users rate at the end of the dialog, is too expensive to collect and can be unreliable due to the existence of malicious users~\cite{horton2016attack}.
While user simulators are proposed to replace real users, they require tons of human work to build and suffer from the discrepancy with real users. 
To cope with this problem, we propose to improve MADPL with the logged user feedback received for each system response during online interactions, such as explicit turndowns (e.g., ``No, I asked for Chinese food!'') and implicit behaviors of retries and interruptions (see \cref{fig:casefig}).
The logged positive and negative user feedback, known as ``bandit feedback'', is cost-efficient to collect and faithful to real-world scenarios, and has been widely exploited in various applications such as machine translation~\cite{kreutzer2018can} and spoken language understanding~\cite{falke2021feedback}.

MADPL from logged user feedback is challenging since the feedback provides only partial information limited to the particular dialog actions predicted during historical interaction.
The feedback for all the other predictions that the system could have made is typically not known.
Moreover, the logged predictions are influenced by the historical policy deployed in the system (i.e., the logging policy), which may over or under-represent specific actions, yielding a biased data distribution~\cite{swaminathan2015batch,gao2021enhancing}. 
Some straightforward ways could be derived from direct method (DM) or counterfactual risk minimization (CRM) algorithms while suffer from {\em extrapolation error}, which is analogous to offline RL~\cite{jaques2020human} (refer to Section~\ref{sec:related} for detailed discussions). 
Specifically, task-oriented dialogs share the ``one-to-many'' nature that various appropriate responses may exist under the same context~\cite{huang2020semi}.
Accordingly, in the multi-action setting, various action combinations could be considered appropriate under the same context, which are mostly not explored by the logging policy.
With the biased and insufficient logged dataset, existing methods will learn arbitrarily bad estimates of the reward of the unexplored region of dialog state-action space.
Thus, when facing real-world scenarios, they can generate noisy dialog actions under-explored during training and result in redundant conversations.

\begin{figure*}[tp]
  \includegraphics[width=\linewidth]{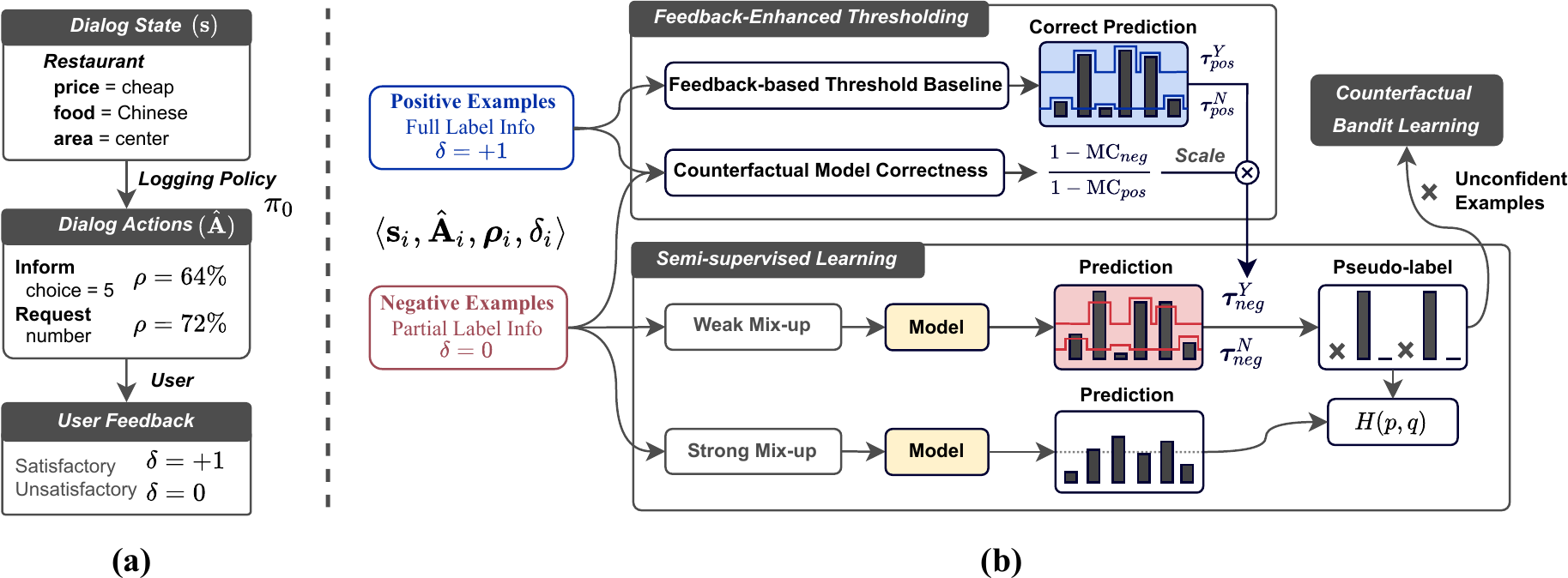}
  \centering
  \caption{\textbf{(a) Data logging procedure}.
    During online interaction, an online agent $\pi_0$ (i.e., logging policy)
    responds to the dialog state $\bs$ that summarizes the dialog history with
    multiple dialog actions $\hat{\bA}$.
    We collect $(\bs, \hat{\bA})$ together with the corresponding prediction
    probability $\brho$ and user feedback $\delta$ as logged bandit data.
    \textbf{(b) Diagram of BanditMatch.}
    The user feedback examples are first fed into the current policy to learn the
    feedback-enhanced thresholds.
    Specifically, the threshold baselines are set based on the average of
    probabilities of correctly predicted classes in positive examples.
    The feedback-enhanced thresholds for the negative examples are derived by
    scaling the baselines by comparing the model correctness for negative and
    positive examples.
    Then, BanditMatch partitions the negative examples into confident and
    unconfident ones based on the thresholds.
    Finally, the positive examples and confident negative examples are leveraged by
    SSL to learn from full label feedback while the rest unconfident ones by
    bandit learning to fully exploit the feedback information and, in turn, enhance
    pseudo-labeling.}
  \label{fig:framework}
\end{figure*}

In this work, we cast the task of MADPL from logged user feedback as a feedback-enhanced semi-supervised learning problem and propose a method BanditMatch to solve this problem.
\zhang{BanditMatch integrates pseudo-labeling methods to better explore the action space through constructing full label feedback.
Specifically, the logged dataset is partitioned into two subsets of full feedback and partial feedback examples, corresponding to positive and negative user feedback. The partial feedback examples with high-confidence predictions are assigned with a pseudo label to learn from full label feedback, while only the rest unconfident ones from partial feedback.
We further propose a feedback-enhanced thresholding (FET) strategy that derives dynamically adjusted adaptive thresholds based on the model performance evaluated with the logged examples.
Unlike fixed pre-defined thresholds, FET resists ignoring the partial feedback examples with correct pseudo labels or selecting those with wrong ones.}
A diagram of BanditMatch is shown in Fig.~\ref{fig:framework}.

Our main contributions are summarized as follows:
\begin{itemize}[leftmargin=*]
\item
We explore to improve MADPL with logged bandit feedback.
Compared with fully labeled expert examples and sparse real user ratings, bandit feedback is cost-efficient to collect and faithful to real-world scenarios. 
\item
We propose BanditMatch to solve MADPL from logged user feedback from a semi-supervised learning perspective. 
Our method can effectively increase the generalization ability while alleviating the extrapolation error caused by limited exploration, and has a broader positive impact on other feedback-related applications.
\item
\zhang{We conduct extensive experiments on the benchmark MultiWOZ dataset,
and empirical results show that our BanditMatch
achieves state-of-the-art performance in task success rate while generating much more concise and informative responses ($9\% \sim 23\%$ increase in Inform F1).}
\end{itemize}

\section{Related work}
\noindent
\textbf{Multi-Action Dialog Policy Learning (MADPL).}
Recent efforts have been made to learn a multi-action dialog policy for task-oriented dialogs where the agent can produce multiple actions per turn to expand its expressive power.
Early attempts~\cite{gordon2020learning} consider the top-$k$ most frequent dialog action combinations in the dataset and simplify MADPL as a multi-class classification problem.
However, the expressive power and flexibility of the dialog agent are severely limited. 
Recent works~\cite{shu2019modeling,li2020rethinking} typically cast MADPL as a multi-label classification problem and address it with supervised learning.
However, the complex action combinations can
exponentially enlarge the output space~\cite{jhunjhunwala2020multi}.
The limited coverage of the output action space of the existing dialog corpus will greatly hinder SL-based methods' performance. 
\citet{jhunjhunwala2020multi}) addresses this problem by data augmentation through interactive learning.
The problem can also be partially alleviated through RL-based and adversarial learning based
methods~\cite{takanobu2019guided,li2020rethinking} by allowing the model to explore more dialog scenarios.
Unlike existing works that rely on limited training corpus or unstable and sparse rating rewards from real users or user simulators, we explore to improve MADPL with logged bandit feedback that is cost-efficient to collect and faithful to real-world scenarios. 

\vskip 6pt
\noindent
\textbf{Learning from Logged User Feedback.}
The user feedback logged during online interaction (e.g., the number of ranked articles the user read), known as ``bandit feedback'', can be recorded at little cost and in huge quantities~\cite{joachims2018deep}.
Learning from bandit feedback is typically addressed from two lines of algorithms: direct method (DM) and counterfactual risk minimization (CRM).
DM learns to directly predict user rewards with a regression estimator and takes as a prediction the action with the highest predicted reward~\cite{beygelzimer2009offset}.
However, the estimator can be misspecified by the partial observability of the environment~\cite{sachdeva2020off}.
To address these issues, domain adaptation methods are proposed to improve the regression model, where the source domain is observational data and the target domain is a randomized trial~\cite{johansson2016learning,gao2021enhancing}.
Despite their success, DMs do not directly apply to our multi-label classification setting as they ignore inter-label dependencies and can fail to generate action combinations with the arg-max operation.

CRM, also known as batch learning from bandit feedback, optimizes the policy model by maximizing its reward estimated with a counterfactual risk estimator~\cite{dudik2011doubly,swaminathan2015batch,johansson2016learning,joachims2018deep,swaminathan2017off}, while requires the strong assumption that the logging policy has full support for every policy in the policy space.
With deficient support common in real-world systems, vanilla methods suffer from extrapolation error, learning arbitrarily bad estimates of the value of responses not contained in the data~\cite{jaques2020human}.
To address this problem, \citet{sachdeva2020off} proposes regression extrapolation that estimates the reward for under-explored examples with a learned model. Just as DM methods, it does not directly apply to multi-label classification.
Policy space restriction methods are derived from offline RL, restricting the model from staying close to its own logging policy through KL-control~\cite{jaques2020human,sachdeva2020off,zhang2020contract}.
However, KL-control depends on the logging policy quality and does not work well for low-quality ones. In comparison, we address the extrapolation error with pseudo-labeling that improves the logging policy step-by-step. It works for low-quality logging policies and alleviates the cold start problem.

\vskip 6pt
\noindent
\textbf{Semi-Supervised Learning.}
Semi-supervised learning is a mature field with a vast diversity of approaches. We focus on pseudo-labeling methods that use hard artificial labels converted from model predictions, and refer the reader to \cite{yang2021survey} for broader introductions.
Specifically, the confidence-based thresholding, which uses unlabeled data only when the predictions are sufficiently confident, is presented in state-of-the-art methods UDA~\cite{xie2020unsupervised} and FixMatch~\cite{sohn2020fixmatch}, where a fixed threshold is chosen to determine confident examples.
In addition, several works have investigated the dynamic thresholds~\cite{xu2021dash,zhang2021flexmatch}. In our work, it is the first attempt to learn an adaptive confidence threshold from the user feedback data. Besides, the unconfident examples are also leveraged to improve the model through counterfactual bandit learning and, in turn, enhance pseudo-labeling.\label{sec:related}

\section{Problem Formulation}
We regard MADPL as a multi-label classification task.
Given dialog state $\bs_i$, we want to predict a macro system action $\bA_i$ that
contains several atomic dialog actions.
Following~\cite{li2020rethinking}, each atomic dialog action is a concatenation of
domain name, action type, and slot name, e.g., “hotel-inform-area”.

Unlike supervised learning with expert examples $(\bs_i, \bA_i)$, we want to
improve MADPL with logged user feedback $(\bs_i, \hat{\bA}_i, \brho_i, \delta_i)$.
$\hat{\bA}_i$ is the historical prediction made by a logging policy $\pi_0$ for
input $\bs_i$ with probabilities $\brho_i=\pi_0(\hat{\bA}_i\mid \bs_i)$, and
$\delta_i \in \{0,+1\}$ is the user feedback received for that prediction.
See \cref{fig:framework} for the data logging procedure.
We denote the data with positive and negative feedback as {\em positive examples}
and {\em negative examples}, respectively.
Note that the logged user feedback provides only partial information limited to
the actions taken by $\pi_0$ and excludes all other actions the system could have
taken.
While positive examples can be viewed as fully labeled ones as in traditional
supervised learning, the ground-truth labels for negative examples are typically
unknown.

\section{Algorithm}

We propose BanditMatch to address the MADPL task.
BanditMatch consists of three modules, i.e., the {\em feedback-enhanced
  thresholding} (FET) module, the {\em semi-supervised learning} (SSL) module, and
the {\em counterfactual bandit learning} (CBL) module, as shown in
\cref{fig:framework}.
Unlike straightforward bandit learning from partial label feedback for all the
negative examples, we propose to learn from full label feedback for the negative
examples with high-confident predictions through pseudo-labeling.
Specifically, the bandit data is partitioned into three parts: fully labeled positive examples
$D_L$, pseudo-labeled confident
negative examples $D_P$, and partially labeled unconfident negative
examples $D_U$.
$D_P$ and $D_U$ are split by comparing current prediction
probability with the adaptive decision boundaries learned by FET.
We further propose a hybrid learning objective to fully explore all the available
label feedback for updating model parameters, where the positive examples $D_L$
and negative confident examples $D_P$ are used to learn through SSL, while the rest
negative unconfident examples $D_U$ through CBL.


\subsection{Feedback-Enhanced Thresholding}

Recent SSL algorithms~\cite{sohn2020fixmatch,xie2020unsupervised} select
high-confidence unlabeled data with a fixed confidence threshold for each class.
However, the learning status varies for different dialog scenarios due to their various learning difficulty. A fixed confidence threshold that treats all examples equally can be an over-high standard for under-learned examples and an over-low standard for over-learned ones. This may lead to eliminating examples with correct pseudo labels and selecting those wrong ones.
Noticing this weakness, we propose the {\em feedback-enhanced thresholding} (FET)
module to evaluate model performance with the feedback data and adjust the
confidence threshold for each individual class at each single training step.

Ideally, we want to obtain negative examples correctly predicted by the updated policy $\pi$, and compute the threshold on the negative examples by averaging the corresponding confidence scores.
Though appealing, this is impractical since the ground-truth labels for negative
examples are unavailable.
To address this issue, we propose to estimate such average confidence scores with correctly predicted positive examples.
Intuitively, the better the model performs, the more credible its predictions are, and the lower the confidence threshold that can be set for it. Considering an extreme case, we can set a low threshold (e.g., 50\%) for an always-correct oracle policy because its predictions are 100\% reliable. In other words, the confidence thresholds for specific examples are negatively related to the correctness of the policy on those examples.
Therefore, taking thresholds on correctly predicted positive examples $\btau_{pos}$ as baselines,
we propose to obtain the thresholds on negative examples $\btau_{neg}$ by scaling the baselines with model correctness $\text{MC}(\cdot)$, as:
\begin{align*}
  \btau^{Y}_{neg} &\triangleq \btau^{Y}_{pos} \frac{1 - \text{MC}_{neg}(\pi)}{1 - \text{MC}_{pos}(\pi)} \\
  \btau^{N}_{neg} &\triangleq \vec{1} - (\vec{1} - \btau^{N}_{pos}) \frac{1 - \text{MC}_{neg}(\pi)}{1 - \text{MC}_{pos}(\pi)}
\end{align*}
where $\vec{1}$ denotes the one-vector.
Note we address a multi-label classification problem where class presence is independent so both positive and negative labels are necessary for training. Thus, two sets of thresholds $\btau^{Y(N)}_{*} = (\tau^{Y(N)}_{*,1}, \ldots,
\tau^{Y(N)}_{*,C})$ are used to select (ban) a specific dialog action $a_c$.

Specifically, we first apply the updated policy $\pi$ on positive examples $D_L$ and obtain the correctly predicted ones $D_{T} = \{(\bs_i, \hat{\bA}_i), i=1,\ldots,N_T\}$.
The thresholds for positive examples are then calculated by:
\begin{align*}
  \tau^{Y}_{pos,c} &\triangleq \frac{1}{N^{gc}_T} \sum_{i=1}^{N_T}{\mathbbm{1}(a_c \in \hat{\bA}_i) \pi(a_c | \bs_i)} \\
  \tau^{N}_{pos,c} &\triangleq \frac{1}{N^{lc}_T} \sum_{i=1}^{N_T}{\mathbbm{1}(a_c \notin \hat{\bA}_i)\pi(a_c | \bs_i)}
\end{align*}
where $N^{gc}_T$ (resp.~$N^{lc}_T$) is the number of correct predictions that
contain (resp.
exclude) the $c$-th dialog action, $\mathbbm{1}(\cdot)$ is the indicator function.
We further construct the model correctness with its normalized expected feedback
estimated from the bandit data, i.e.,
\resizebox{\linewidth}{!}{
\begin{minipage}{\linewidth}
\begin{align*}
  \text{MC}_{pos}(\pi) &\triangleq \frac{1}{| D_T| }\sum_{x_i\in D_T}
  \sum_{\hat{a}_j\in\hat{\bA}_i}\text{Attr}_{pos}(\hat{a}_j, \hat{\bA}_i)\frac{\pi(\hat{a}_j| \bs_i)}{\pi_0(\hat{a}_j| \bs_i)}\\
  \text{MC}_{neg}(\pi) &\triangleq \frac{1}{| B_N| }\sum_{x_i\in B_N}\sum_{\hat{a}_j\in\hat{\bA}_i}\text{Attr}_{neg}(\hat{a}_j, \hat{\bA}_i)\frac{1 - \pi(\hat{a}_j| \bs_i)}{1 - \pi_0(\hat{a}_j| \bs_i)}
\end{align*}
\end{minipage}
}

where $B_N$ are the negative examples in the current batch.
We use attribution functions to determine which individual dialog action caused
the error in the predicted $\hat{\bA}_i$.
Specifically, for positive examples we simply average over the predicted dialog
actions as they contribute equally to the success, i.e., $\text{Attr}_{pos}(\cdot, \hat{\bA}_i)\triangleq1/|\hat{\bA}_i|$.
For negative examples, we follow the idea that the policy might be self-aware of
some mistakes and distributes the feedback proportional to the uncertainty
reflected in the propensities~\cite{falke2021feedback}, i.e.,
\begin{align*}
  \text{Attr}_{neg}(\hat{a}_j, \hat{\bA}_i) \triangleq \frac{\pi_{0}(\hat{a}_j| \bs_i)}{\sum_{a^*_k\in\hat{\bA}_i}\pi_{0}(a^*_k| \bs_i)}
\end{align*}

\subsection{Feedback-Enhanced Semi-Supervised Multi-Action Dialog Policy Learning}
Given the feedback-enhanced thresholds, we are now able to perform semi-supervised learning and learn from full label information with the fully labeled positive examples $D_L$ and pseudo-labeled confident negative examples $D_P$.
To make the most of all the available label information, we propose to improve the pseudo-labeling model with the rest unconfident negative examples $D_U$ through bandit learning with CRM.
A KL control method is further applied to prevent the policy from choosing unrealistic dialog action combinations.
Unlike prior SSL algorithms~\cite{sohn2020fixmatch,zhang2021flexmatch}, we do not pre-split the data examples but determine them in each batch adaptively during training.

\subsubsection{Semi-supervised Learning}
\zhang{We select FixMatch~\cite{sohn2020fixmatch}, a simple yet effective SSL example, as the base of the proposed SSL algorithm.
FixMatch first generates pseudo-labels using the model’s predictions on weakly augmented unlabeled examples, where a fixed threshold is set to select high-confidence predictions.
The model is then trained to predict the pseudo-label when fed a strongly-augmented version of the same example.
For our SSL algorithm, we extend FixMatch to a multi-label classification setting with dialog states as training examples and apply the feedback-enhanced thresholds instead of fixed ones.}

Specifically, following FixMatch, our SSL consists of two binary cross-entropy loss terms $\mathcal{L}_L$ and $\mathcal{L}_P$, which are applied to positive examples and confident negative examples, respectively.
Specifically, $\mathcal{L}_L$ is the standard binary cross-entropy loss on weakly augmented labeled examples, as:
\begin{align*}
\mathcal{L}_L = \frac{\sum_{i=1}^{B}\mathbbm{1}(\delta_i = 1)H(\hat{\bA}_i, \pi(\bA\mid\omega(\bs_i, \alpha_w)))}{\sum_{i=1}^{B}\mathbbm{1}(\delta_i = 1)}
\end{align*}
where $B$ is the batch size, and $\omega$ performs mix-up operation on $\bs_i$.
Unlike pictures or natural language sentences, dialog states are usually represented with a dictionary object and encoded with one-hot vectors~\cite{takanobu2019guided,li2020rethinking}.
We thus consider mix-up~\cite{zhang2017mixup} as datatype-agnostic data augmentation for consistency regularization.
It mixes the training examples as:
\begin{align*}
\bs_i = \lambda \bs_i + (1 - \lambda) \bs_j
\end{align*}
where $\lambda$ is sampled from a Beta distribution, i.e., $\lambda = \text{max}(\mathtt{Beta}(\alpha_w, \alpha_w), 1-\mathtt{Beta}(\alpha_w, \alpha_w))$, and $\alpha_w$ is the shape parameter of the distribution for weak augmentation.

Unlike the multi-class classification setting in~\cite{sohn2020fixmatch} that accepts or rejects a whole example, the feedback-enhanced thresholds filter a set of confidently accepted (rejected) atomic dialog actions (i.e., confident classes) for each example.
By viewing multi-label classification with $|\bA|$ classes as $|\bA|$ binary classifications, we obtain a pseudo label for the $c$-th class with the $i$-th weakly-augmented example as $\hat{q}_{i,c} = \mathbbm{1}(\pi(a_c\mid\omega(\bs_i, \alpha_w)) > 0.5)$, and enforce the cross-entropy loss on confident classes as:
\begin{align*}
\mathcal{L}_P = \frac{\sum_{i=1}^{B}\sum_{c=1}^{C}\text{Conf}_{i,c} H(\hat{q}_{i,c}, \pi(a_c\mid\omega(\bs_i, \alpha_s)) )}{\sum_{i=1}^{B}\sum_{c=1}^{C}\text{Conf}_{i,c}}
\end{align*}
where $\alpha_s$ is the shape parameter of the Beta distribution for strong augmentation.
The confidence indicator is calculated by comparing the prediction of weakly augmented example and the feedback enhanced thresholds, i.e., $\text{Conf}_{i,c} = \mathbbm{1}(\delta_i = 0 \wedge \pi(a_c \mid \omega(\bs_i, \alpha_w)) \notin [\tau_{neg,c}^{N},\tau_{neg,c}^{Y}])$.

\subsubsection{Counterfactual Bandit Learning}
For the unconfident examples, the cross-entropy loss cannot be used to guide the training process since the predictions can be erroneous.
We thus apply CRM to improve the pseudo-labeling model through counterfactual bandit learning.
The goal of learning is to find a policy $\pi$ that maximizes the expected payoff, as:
\begin{align*}
R(\pi) =  \mathbb{E}_{\bs\sim\mathtt{Pr}(S)}\mathbb{E}_{\bA\sim\mathtt{Pr}(\mathbb{A}\mid \bs)}[\delta(\bs,\bA)]
\end{align*}

Specifically, we aim to minimize the negative expected feedback estimated with the pseudoinverse estimator (PI)~\cite{swaminathan2017off}, defined as:
\begin{align*}
\mathcal{L}_B = - \frac{\sum_{i=1}^{B} \delta_i \cdot (1 + \sum_{c=1}^{C} \text{Unconf}^+_{i,c} \cdot (\frac{\pi(a_c\mid \bs_i)}{\rho_{i,c}}-1))}{\sum_{i=1}^{B}\sum_{c=1}^{C}\text{Unconf}^+_{i,c}}
\end{align*}
where $\text{Unconf}^+_{i,c}$ is an indicator function that filters positive examples besides unconfident negative ones to avoid a degenerate training objective~\cite{swaminathan2015batch}.
Notably, the bandit learning process can boost model performance in the early training stage when confident examples are rare and effectively improve pseudo-labeling.

\subsubsection{KL Control from the Logging Policy}
The joint bandit learning and semi-supervised learning allow a controlled exploration of state-action space. However, it still can generate unrealistic action combinations.
To address this issue, we directly incorporate knowledge of the logging policy, which imitates expert dialog action combinations, into the training process through KL control~\cite{jaques2020human}, i.e.,
\begin{align*}
\mathcal{L}_K = D_{KL}[\pi(\bs)||\pi_0(\bs)]
\end{align*}
where $D_{KL}[q||p] = \sum_x q(x)(\log{q(x)}-\log{p(x)})$ calculates the KL divergence between probability distributions.

\subsubsection{Overall Objective Function}
Our proposed BanditMatch algorithm optimizes end-to-end with the overall loss as the weighted sum of four losses:
\begin{align*}
\mathcal{L} = \mathcal{L_L} + \lambda_p\mathcal{L_P} +  \lambda_b\mathcal{L_B} + \lambda_k\mathcal{L_K} 
\end{align*}
where $\lambda_p$, $\lambda_b$, and $\lambda_k$ are hyper-parameters to balance each term’s intensity. We find that simply setting them to 1 can lead to proper performance through experimental evaluation.


\section{Experiments}\label{sec:exp}
\begin{table*}[tp]
\centering
\caption{Interactive evaluation results with $10\%$ annotated examples. We simulate 1,000 dialogs per run and report the mean and standard deviation over $5$ runs. We focus on comparing Inform F1 (low Inform F1 indicates dialog redundancy) and Success.}
\label{tab: robust}
\resizebox{.7\textwidth}{!}{
\small
\begin{tabular}{llccccc}
\toprule
Category & Agent & Turn & Match & Inform Rec & \textbf{Inform F1}  & \textbf{Success\%} \\
\midrule
\multirow{2}*{SL} & Full SL  & 11.46 \small{$\pm 0.56$} & 0.68 \small{$\pm 3.9\%$} & 0.81 \small{$\pm 3.2\%$} & \textbf{0.81} \small{$\pm 2.1\%$} & 67.3 \small{$\pm 3.69$} \\
\cmidrule{2-7}
& Logging Policy & 13.65 \small{$\pm 0.72$} & 0.61 \small{$\pm 7.1\%$} & 0.73 \small{$\pm 4.9\%$} & 0.79 \small{$\pm 3.2\%$} & 54.3 \small{$\pm 5.47$} \\
\midrule
\multirow{3}*{RL} &
PPO \footnotemark[1] & 12.9 & - & - & - & 33.8 \\
& ALDM\footnotemark[1] & 14.9 & - & - & - & 36.4 \\
& GDPL\footnotemark[1] & 11.8 & - & - & - & 50.2 \\
\midrule
\multirow{2}*{SSL} &
FixMatch & 14.01 \small{$\pm 2.89$} & 0.56 \small{$\pm 15.8\%$} & 0.68 \small{$\pm 18.3\%$} & 0.69 \small{$\pm 8.5\%$} & 50.3 \small{$\pm 19.3$} \\
&Act-VRNN \footnotemark[1] & \textbf{8.4} & - & - & - & 72.4 \\
\midrule
\multirow{2}*{CRM} &
IPS & 13.60 \small{$\pm 0.14$} & 0.79 \small{$\pm 0.4\%$} & 0.84 \small{$\pm 0.4\%$} & 0.6 \small{$\pm 0.3\%$} & 71.3 \small{$\pm 0.89$} \\
&+ KL control & 11.86 \small{$\pm 0.09$} & 0.80 \small{$\pm 1.0\%$} & \textbf{0.89} \small{$\pm 0.5\%$} & 0.67 \small{$\pm 0.5\%$} & 75.5 \small{$\pm 0.95$} \\
&BanditNet & 14.37 \small{$\pm 0.07$} & 0.8 \small{$\pm 0.3\%$} & 0.87 \small{$\pm 0.4\%$} & 0.53 \small{$\pm 0.2\%$} & 74.3 \small{$\pm 0.54$} \\
&+ KL control & 13.10 \small{$\pm 0.06$} & \textbf{0.83} \small{$\pm 0.3\%$} & 0.87 \small{$\pm 0.2\%$} & 0.6 \small{$\pm 0.2\%$} & 75.8 \small{$\pm 0.04$} \\
\midrule
\multirow{1}*{SSL+CRM}
&\textbf{BanditMatch} & 10.32 \small{$\pm 0.71$} & 0.79 \small{$\pm 2.94\%$} & \textbf{0.89} \small{$\pm 2.1\%$} & 0.76 \small{$\pm 2.2\%$} & \textbf{76.7} \small{$\pm 2.83$} \\
&- MC scale & 10.45 \small{$\pm 0.44$} & 0.76 \small{$\pm 4.5\%$} & 0.88 \small{$\pm 1.8\%$} & 0.75 \small{$\pm 2.2\%$} & 75.6 \small{$\pm 2.36$} \\
&- FET & 10.64 \small{$\pm 0.29$} & 0.77 \small{$\pm 6.9\%$} & 0.88 \small{$\pm 2.8\%$} & 0.73 \small{$\pm 2.1\%$} & 74.4 \small{$\pm 4.76$} \\
&- CBL & 11.54 \small{$\pm 1.68$} & 0.69 \small{$\pm 10.9\%$} & 0.81 \small{$\pm 8.5\%$} & 0.79  \small{$\pm 2.0\%$} & 65.9 \small{$\pm 10.7$} \\
&- KL control & 10.2 \small{$\pm 0.39$} & 0.76 \small{$\pm 1.7\%$} & 0.87 \small{$\pm 3.0\%$} & 0.72  \small{$\pm 1.8\%$} & 73.1 \small{$\pm 2.7$} \\
&- all & 11.5 \small{$\pm 1.02$} & 0.69 \small{$\pm 5.2\%$} & 0.81 \small{$\pm 3.8\%$} & 0.78  \small{$\pm 1.1\%$} & 64.5 \small{$\pm 4.3$} \\
\bottomrule
\end{tabular}
}
\end{table*}

\subsection{Settings}
\zhang{\textbf{Data}~~We use MultiWOZ 2.0~\cite{budzianowski2018multiwoz}, a large-scale multi-domain benchmark dataset, to validate the effectiveness of our method and apply the agenda-based user simulator as the interaction environment to evaluate the generalization ability. Other datasets
are not considered since the corresponding user simulators are unavailable.}

\zhang{To simulate MADPL from logged user feedback with the above dataset, we follow~\citet{falke2021feedback} and consider a simplified scenario where the user always provides correct feedback.
The user feedback $\delta_i$ is determined by comparing the logging policy prediction and the ground truth.
Specifically, we split the train set into $p\%$ labeled data ($D_S$), on which the logging policy is trained with supervised learning, and use the rest to create bandit data ($D_B$).
For each state $s_i$ in $D_B$, we follow the multi-label classification setting and take as agent prediction the actions whose prediction probability is greater than $0.5$~\cite{rizve2021defense}.
The user feedback $\delta_i = 1 (0)$ if the predicted labels do (not) match the ground truth.
Note that $\delta_i$ is set to $0$ rather than $-1$ for negative feedback to avoid a degenerate training objective~\cite{swaminathan2015batch}.}

\zhang{In real-world scenarios, the user feedback can be captured by an extra module that detects user behaviors such as paraphrasing, retries, etc~\cite{falke2020leveraging}.
The corresponding last system responses are then determined as positive (negative) if there are no (any) detected negative behaviors.
The feedback can be wrong due to the inaccuracy of the detection and the randomness of user behaviors.
Such a noisy user feedback setting has been explored by~\citet{agarwal2021learning}, where the idea of unbiased estimation of the feedback can be further integrated to improve BanditMatch.}

\vskip 6pt
\noindent
\textbf{Baselines}~~
We compare BanditMatch with state-of-the-arts: 1) RL methods: PPO~\cite{schulman2017proximal}, ALDM~\cite{liu2018adversarial}, and GDPL~\cite{takanobu2019guided}; 2) CRM methods: IPS~\cite{swaminathan2015batch} and BanditNet~\cite{joachims2018deep}; and 3) SSL methods: Act-VRNN~\cite{huang2020semi} and FixMatch~\cite{sohn2020fixmatch}, under the fully SL-based skyline~\cite{li2020rethinking}. 

\zhang{The SL-based skyline is trained on all the trainset examples. Other models are further fine-tuned based on the logging policy. Specifically, the RL methods are fine-tuned with the same agenda-based user simulator for evaluation, while the rest methods with the bandit data $D_B$.}

\vskip 6pt
\noindent
\zhang{\textbf{Evaluation Metrics}~~
We perform the automatic evaluation on task completion quality through dialog interaction with the user simulator.
We use {\em Match}, {\em Turn}, {\em Success}, {\em Inform Recall}, and {\em Inform F1} as evaluation metrics:
{\em Match} evaluates whether the booked entities match the indicated constraints (e.g., the cheapest restaurant in town).
{\em Inform Recall} and {\em Inform F1} evaluate whether all the requested information (e.g., hotel address) has been provided.
{\em Turn} and {\em Success} evaluate the efficiency and level of task completion of dialog agents, respectively.
Specifically, a dialog is successful if all the requested information is provided (i.e., {\em Inform Recall}~$= 1$) and all the entities are correctly booked (i.e., {\em Match}~$= 1$).}

\zhang{Note that the partial user feedback information can result in successful but highly redundant conversations (see Introduction). For example, the agent may inform the user about tons of details about a hotel when the user only asks for its address. Such redundancy is reflected by {\em Inform F1}, which we thus consider as a critical metric besides {\em Success}.}

\footnotetext[1]{Results reused from~\cite{huang2020semi}}

\subsection{Algorithm Evaluation}
\noindent
\textbf{Overall Comparisons.}
Table~\ref{tab: robust} shows the performance of different methods.
\zhang{We observe that our method achieves the highest performance in the task success by $0.9\%\sim5.4\%$ while achieving a significant $9\%\sim23\%$ increase in Inform F1 score.}
This demonstrates that BanditMatch can effectively explore more state-action space to resist extrapolation error and generate more concise responses through pseudo-labeling confident examples.
Compared with the SSL-based FixMatch, BanditMatch reaches a substantially better and stable performance in all evaluation metrics on account of the proper use of bandit user feedback.
Besides, BanditMatch generally outperforms the Full SL method (which uses full label information) but fails to reach the same inform F1 score.
This is because SL can over-fit specific dialog action combinations from limited demonstrations and ignore all the other possible valid responses that the agent could have taken, while CRM allows BanditMatch to explore unseen combinations.

\begin{figure*}[htp]
  \includegraphics[width=.85\linewidth]{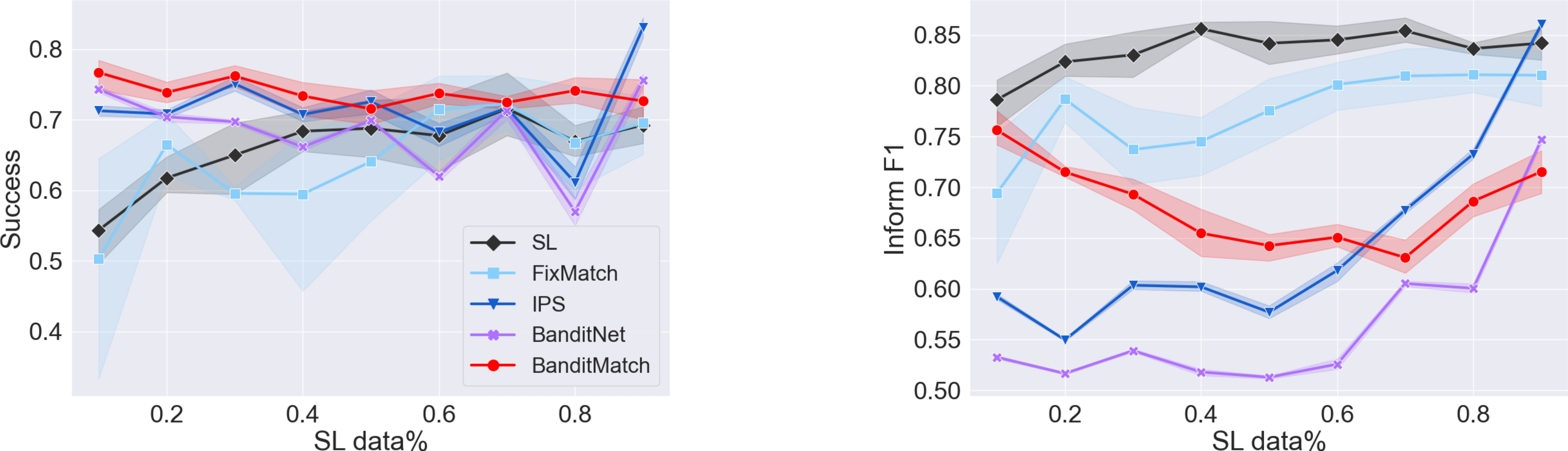}
  \centering
  \caption{Agent performance under different percentage of expert examples (SL data\%).}
  \label{fig:pct}
\end{figure*}

\vskip 6pt
\noindent
\textbf{Ablation Study.}
To gain deeper insight into the contributions of different components involved in our approach, we conduct ablation studies by considering the following variants:
1) a variant without MC scaling and directly uses SL-based minimum thresholds;
2) a variant without FET and applies fixed pre-defined thresholds;
3) a variant without CBL for unconfident examples;
4) a variant without KL control;
5) a variant without all the proposed modules and strategies, i.e., a FixMatch model with additional labeled examples (logged positive examples).
We also report CRM method variants to make a fair comparison over KL control.
The results are merged in Table~\ref{tab: robust}.

We see that BanditMatch consistently outperforms its ablated variants.
Specifically, we find that:
1) The FET strategy (including MC scale) contributes to overall performance gain (especially on Inform F1 score) by increasing the number of confident examples.
2) CBL plays a key role in performance gain and model stabilization as it exploits the feedback from unconfident examples to enhance pseudo-labeling by reducing the confirmation bias.
3) KL control also works by improving Inform F1 score like pseudo-labeling while less effective. In addition, the vanilla combination of CRM and KL control still suffers from severe extrapolation error.

\begin{table}[tp]
\small
\centering
\caption{Human evaluation results (mean over 10 judges).}
\label{tab: human}
\begin{tabular}{lccc}
\toprule
Dialog pair & Win & Lose & Tie \\
\midrule 
Ours vs SL & 56.6 & 23.9 & 19.5 \\
Ours vs BanditNet & 60.3 & 25.9 & 13.8 \\
Ours vs FixMatch & 52.6 & 28.0 & 19.4 \\
\bottomrule
\end{tabular}
\end{table}

\vskip 6pt
\noindent
\textbf{Human Evaluation.}
Automatic evaluation only measures part of the agent’s performance (e.g., Success for the level of task completion). It may fail to consider other aspects for assisting real users (e.g., redundant or inappropriate reply). 
Thus, we conduct a human study to fully evaluate system performance.
Our BanditMatch method is compared with three methods: SL, FixMatch, and BanditNet.
Following prior works~\cite{takanobu2019guided,li2020rethinking}, for each comparison pair, we randomly example $100$ user goals and present the generated dialog examples from the two agents. 
The dialog actions are translated into human-readable utterances
with a rule-based language generation module. 
We show each example to $10$ judges to select the dialogs that provide a better user experience.

In Table~\ref{tab: human} and \ref{tab: case}, we see that our method outperforms the baseline methods. Especially compared to BanditNet, though comparable in success rate, our BanditMatch method provides high-quality service by generating concise and informative system responses.

\vskip 6pt
\noindent
\textbf{Discussion.}
In what follows, we study how many expert examples are enough for proper MADPL from logged user feedback.
Given the MultiWOZ dataset, we build $10$ random splits based on the percentage of expert examples.
The expert examples are used to initiate the logging policy, and the rest to create bandit data.
We report the results in Fig.~\ref{fig:pct}.

We observe that BanditMatch achieves an improved and stable success rate across different SL data percentages, demonstrating our method's effectiveness on MADPL while reducing human annotation costs.
The Inform F1 score first decreases and then goes up with the growth of SL data percentage.
We speculate that the reason is that with the decrease of bandit data, the estimation of FET tends to be biased and increases the percentage of unconfident data, making the performance more influenced by CBL.
Note that we can make a trade-off between conciseness and information richness by adjusting the weight for the CBL loss.
With respect to vanilla CRM methods, we see substantially better inform F1 score in the high SL percentage region.
In other words, the main contributions come from the SL stage, while they can not achieve the same performance without a favorable logging policy.
\begin{table}[tp]
\small
\centering
\caption{Case system response. We see that the bandit learning method suffers from severe redundancy, while our BanditMatch generates concise and informative response.}
\label{tab: case}
\begin{tabularx}{\linewidth}{X}
\textbf{[User]}: I'd like to find an expensive restaurant in the centre.\\
\midrule
\textbf{[BanditNet]}: There are 4 available restaurants. How about ugly duckling? That is an expensive Chinese restaurant in the centre area, located at \ldots phone number as \ldots postcode is \ldots. Would you like to make a reservation? What time would you like the reservation for? Is there anything else I can help you with today?\\
\midrule
\textbf{[Ours]}:  Found 4 such places. What type of food do you like?\\
\end{tabularx}
\end{table}

\section{Conclusion and Future Work}
This paper studies the problem of multi-action dialog policy learning from logged user feedback.
To address this problem, we present a feedback-enhanced semi-supervised algorithm BanditMatch, which complements the partial labels through pseudo-labeling.
Benefiting from fully utilizing all the label information through joint semi-supervised and counterfactual bandit learning, BanditMatch is able to alleviate the extrapolation error and improve model performance towards unseen dialog flows while generating more concise and informative responses.
Extensive experiments demonstrate the superiority of our proposed method.
In the future, to improve data privacy for the user feedback, we plan to explore the federated bandit learning scenario where the personal user feedback is stored locally in each client and not exchanged or transferred.

\section*{Acknowledgments}

This work was supported in part by National Key R\&D Program of China
(2021YFB1715600), National Natural Science Foundation of China
(61902305, 62272372, 61922067), and
MoE-CMCC ``Artifical Intelligence'' Project (MCM20190701).

\bibliography{references}

\end{document}